\documentclass[english]{article}
\usepackage[T1]{fontenc}
\usepackage[latin9]{inputenc}
\usepackage{geometry}
\usepackage{mathrsfs}
\usepackage{amsmath}
\usepackage{amsthm}
\usepackage{amssymb}
\usepackage{graphicx}
\usepackage[authoryear]{natbib}

\makeatletter
\numberwithin{equation}{section}
\numberwithin{figure}{section}
\newcommand{\lyxrightaddress}[1]{
	\par {\raggedleft \begin{tabular}{l}\ignorespaces
	#1
	\end{tabular}
	\vspace{1.4em}
	\par}
}
\newenvironment{lyxcode}
	{\par\begin{list}{}{
		\setlength{\rightmargin}{\leftmargin}
		\setlength{\listparindent}{0pt}
		\raggedright
		\setlength{\itemsep}{0pt}
		\setlength{\parsep}{0pt}
		\normalfont\ttfamily}%
	 \item[]}
	{\end{list}}

\makeatother

\usepackage{babel}
\begin{document}
\title{Convolutional \emph{unitary} or \emph{orthogonal} recurrent neural
networks. }
\author{Marcelo O. Magnasco}
\maketitle

\lyxrightaddress{Lab of Integrative Neuroscience, Rockefeller University}
\begin{abstract}
Recurrent neural networks are extremely powerful yet hard to train.
One of their issues is the \emph{vanishing gradient problem}, whereby
propagation of training signals may be exponentially attenuated, freezing
training. Use of \emph{orthogonal }or\emph{ unitary }matrices, whose
powers neither explode nor decay, has been proposed to mitigate this
issue, but their computational expense has hindered their use. Here
we show that in the specific case of convolutional RNNs, we can define
a \emph{convolutional exponential }and that this operation transforms
antisymmetric or anti-Hermitian convolution kernels into \emph{orthogonal}
or \emph{unitary convolution kernels}. We explicitly derive FFT-based
algorithms to compute the kernels and their derivatives. The computational
complexity of parametrizing this subspace of orthogonal transformations
is thus the same as the networks' iteration. 
\end{abstract}

\section{tl;dr}

This is an extremely terse synopsis for quick reference. All proofs,
explanations and misguided attempts at clarity are in subsequent chapters,
to which you're welcome to skip to. 

Given a layer $X$ in $D$ dimensions, a spatial convolution operation
$\otimes$ and a convolution kernel $K$ acting on $X$ as $K\otimes X$,
we formally define the \emph{convolutional exponential} $e_{\otimes}^{K}$
as the kernel defined by the series 
\begin{equation}
e_{\otimes}^{K}\otimes X\equiv X+K\otimes X+\frac{1}{2!}K\otimes K\otimes X+\frac{1}{3!}K\otimes K\otimes K\otimes X+\frac{1}{4!}K\otimes K\otimes K\otimes K\otimes X+\cdots\label{eq:convex_formal}
\end{equation}
so the linear operator defined by $e_{\otimes}^{K}\otimes$ is \emph{quite
literally }the matrix exponential of the linear operator defined by
$K\otimes$. This exponential can be computed in Fourier space through
\begin{equation}
e_{\otimes}^{K}\equiv\mathscr{F}^{-1}\left[\exp(\mathscr{F}\left[K\right])\right]\label{eq:convex_fft}
\end{equation}
where the right-hand exponential is element-wise and where $\mathscr{F}$
and $\mathscr{F}^{-1}$ are the forward and inverse Fourier transforms
in $D$ dimensions, and hence an $N\log N$ operation. This can easily
be generalized to any operation defined through convergent power series,
for example the\emph{ convolutional sine and cosine} of $K$ are defined
through 
\begin{equation}
\begin{array}{l}
\cos_{\otimes}(K)\equiv\mathscr{F}^{-1}\left[\cos(\mathscr{F}\left[K\right])\right]\\
\sin_{\otimes}(K)\equiv\mathscr{F}^{-1}\left[\sin(\mathscr{F}\left[K\right])\right]
\end{array}\label{eq:convcos}
\end{equation}

Given a complex-valued kernel $K$, we define an\emph{ anti-Hermitian
kernel} as one that satisfies $K=-\overline{K^{*}}$ where $\bar{K}$
is the spatial flip operation and $K^{*}$ the elementwise complex
conjugate, because then the linear operator given by $K\otimes$ is
an anti-Hermitian operator. Then $e_{\otimes}^{K}$ is \emph{unitary
}in the sense that the linear operator $e_{\otimes}^{K}\otimes$ is
a \emph{unitary operator}: it is the matrix exponential of the anti-Hermitian
operator $K\otimes$, and as such has eigenspectrum on the unit circle.

Given a complex-valued layer $Z$ in $D$ dimensions, an anti-Hermitian
kernel $K$ acting on $X$, an input $I_{n}$, and element-wise complex-valued
activation function $\phi$, we define a \emph{convolutional unitary
recurrent neural network} (cuRNN?) as the iterated recursion
\begin{equation}
Z_{n+1}=\phi\left(e_{\otimes}^{K}\otimes Z_{n}+I_{n}\right)\label{eq:cuRNN}
\end{equation}
where the subindex $n$ represents the passage of time in the recurrence
and $Z_{0}$ is the initialization value of the layer. 

Given a real-valued layer $X$ in $D$ dimensions, a centrally symmetric
spatial convolution kernel $K$ acting on $X$, an input $I_{n}$,
and real-valued scalar activation functions $\phi$ and $\psi$, we
construct an identical copy of $X$ called $P$, and define a \emph{convolutional
orthogonal recurrent neural network} through the iterated recursion: 

\begin{equation}
\begin{array}{rcl}
X_{n+1} & = & \phi(\ +\cos_{\otimes}(K)\otimes X_{n}\ +\ \sin_{\otimes}(K)\otimes P_{n}\ +\ I_{n})\\
P_{n+1} & = & \psi(\ -\sin_{\otimes}(K)\otimes X_{n}\ +\ \cos_{\otimes}(K)\otimes P_{n}\ )
\end{array}\label{eq:coRNN}
\end{equation}
where the sagacious reader will discern in the arrangement of $\sin_{\otimes}$
and $\cos_{\otimes}$ a rotation matrix in the $XP$ space obtained
through unrolling the real and imaginary parts of $e_{\otimes}^{iK}$
for a real symmetric kernel. For \emph{any} spatially symmetric $K$
this specific combination of $\cos_{\otimes}$ and $\sin_{\otimes}$,
when considered as a linear algebra operator acting on the $X\times P$
space, is an \emph{orthogonal matrix}: it is constructed as the matrix
exponential of an antisymmetric (skew-symmetric) matrix derived from
$K$ and, as such, all its eigenvalues lie on the unit circle. 

Unitary and orthogonal matrices were introduced into the theory of
recurrent networks, for example, in {[}Arjovsky{]} and {[}Vorontsov{]},
to solve the \emph{exploding/vanishing gradient problem} {[}Hochreiter,Pascanu{]},
but their use in practice has remained a challenge because of the
computational cost of maintaining orthogonality during training, either
by re-orthogonalizing, or by exponentiating antisymmetric or antiHermitian
matrices {[}Cardoso{]}. For a convolutional network {[}LeCun1,2,3{]},
Eqs. \ref{eq:convex_fft} and \ref{eq:convcos} give an explicit $N\ln N$
algorithm for transforming a convolution kernel $K$ into a unitary
or orthogonal operation, and thereby solves the computational cost
problem for this specific architecture. 

The remainder of this Paper is as follows. Section 2 is entirely background
on: the exploding/vanishing gradient problem, matrix exponentials
of antisymmetric matrices, the expression of a convolution as a linear
operator, and other needed prolegomena. Section 3 we derive the convolutional
exponential, Section 4 we use it to generate unitary kernels, Section
5 we obtain derivatives of said kernels, and in sections 6,7,8 we
map the complex-valued unitary network, using a bipartite-graph architecture,
to generate orthogonal convolutions. I relegate to SuppMat a more
detailed analysis of the relationship between the full interaction
matrix and a convolution in respect to the exponential operation.
One of the enduring attractions of recurrent NN vs. feed-forward NNs
is that they have been proven to be Turing universal {[}Siegelmann,Kilian{]}.
I am unaware of an extant proof that permits use of the far smaller
space of convolutional RNNs, so I append in the SuppMat a simple sketch
of a proof that convolutional RNNs are TU by embedding a cellular
automaton {[}Cook{]}. The relationship of the convolutional exponentials
to the eigenspectrum of the full linear operator is extremely complex
and beyond our current scope. 

\section{Background}

This section contains background material from a number of different
areas, collected together here for convenience and ease of reference. 

\subsection{The vanishing gradient}

Artificial neural networks usually embody an architecture in which
the main nonlinearities of behavior happen within individual units,
while the propagation of information around the network occurs along
linear connections defined by a synaptic connectivity matrix. A recurrent
neural network is one in which the outputs of the units feed back
on the units themselves and therefore can cause enduring dynamical
activity, usually in the form of 

\begin{equation}
x_{n+1}=\phi(Mx_{n})\label{eq:RNN}
\end{equation}
where $x_{n}$ is a succession of state vectors, $M$ a matrix and
$\phi$ an elementwise nonlinear function usually called the ``activation
function''. RNNs are extremely powerful, in fact Turing universal
{[}Siegelmann,Kilian{]}, and with universality come certain fundamental
limitations. In the theory of dynamical systems, where the emphasis
is not on training but on observing the evolution due to nontrivial
$\phi$, Eq. \ref{eq:RNN} is called a \emph{coupled map lattice}
{[}Kaneko1,2, Alonso{]}. 

The study of the propagation of small perturbations is a centerpost
of dynamical systems theory {[}Strogatz{]}. Perturbations to the activity
feed back onto the system recursively, and may grow or shrink exponentially
as time goes on. Iterated maps like Eq. \ref{eq:RNN} have a telescoping
structure; for instance, if we explicitly iterate it three times we
get
\[
x_{3}=\phi(M\phi(M\phi(Mx_{0})))
\]
and so on and so forth. Such telescoping series of functional compositions
was made for the chain rule, which as applied to this case says that
the long-term fate of perturbations is computed by multiplying together
what happens in individual steps. More specifically, a small perturbation
to the initial condition $x_{0}$ propagates forward using the chain
rule {[}Strogatz{]}
\[
\frac{\partial x_{n}}{\partial x_{0}}=\prod_{i=0}^{n-1}\Lambda_{i}M
\]
where the $\Lambda_{i}$ are diagonal matrices whose elements are
\[
\Lambda_{i}=diag\left(\left.\frac{\partial\phi}{\partial x}\right|_{x_{i}}\right)=diag(\phi'(Mx_{i}))
\]
So the action of each individual timestep is governed by an interaction
between the two network components described: a local amplification
or attenuation due to the activation function, which depends on the
actual values of the states of each neuron, and is given by a diagonal
matrix; and a global amplification or attenuation due to the eigenvalues
of the connectivity matrix, which do \emph{not }depend on the current
state of the neurons or time. There are therefore two contributions
to asymptotic behavior, one being structural and due to the eigenspectrum
of connectivity, and the other one being due to the activation functions.
The structural component grows like $\approx M^{n}$, which, as $n\to\infty$,
is dominated by the eigenvalue of $M$ with the largest absolute value.
For generic matrices, this largest absolute value will not be precisely
equal to $1$ and thus generically this contribution either diverges
or converges to $0$. As we shall see below, a special class of matrices,
the \emph{orthogonal }matrices, have all of their eigenvalues on the
unit circle, and as such their powers do not explode, nor do they
decay, as $n\to\infty$. 

The activation function contribution is more complex, because it is
evolution-dependent. It does have, on the other hand, the advantage
of being diagonal. If $0\le\phi'\le1$ then necessarily there will
be some contraction due to this term, because the slope is never $>1$.
{[}Arjovsky{]} used rectified linear units $\phi(x)=\max(0,x)$ {[}Nair{]},
whose derivative is either $1$ or $0$, to obtain explicit bounds
on behavior. Another possibility is to have a \emph{controlled expansion}
to counter the activation contraction, for instance by using 
\[
\phi(x)=\left(1+\frac{1}{\tau}\right)\max(x,0)
\]
where, since $(1+1/\tau)^{\tau}\to e$ for large $\tau$, we can explicitly
bound the minimum time $\tau$ until a sequence that does not touch
$R^{-}$ reaches a magnification of \emph{$e$}. 

A huge number of methods have been introduced to mitigate the vanishing
gradient problem, such as LSTMs gating, gradient clipping etc. These
methods are outside our scope and have been didactically reviewed
a number of times. 

\subsection{Continuous and discrete time}

Some dynamics are defined on continuous time, through differential
equations, and time is a real number. Other dynamics are defined through
discrete-time iterations, and time is an integer. It is important
to understand how to relate properties of one to the other {[}Strogatz{]}. 

The linear ordinary differential equation 
\[
\dot{x}=Mx
\]
where $x$ is a vector and $M$ a matrix, has explicit solutions obtained
through the matrix exponential 
\[
x(t)=e^{Mt}x(0)
\]
Because the eigenvalues of the matrix exponential are the (scalar)
exponential of the eigenvalues of its argument, the solution either
blows up or exponentially decays as $\approx e^{\lambda t}$ where
the $\lambda$ are the eigenvalues of $M.$ The real part of the $\lambda$
therefore control whether the solutions grow (positive) or decay (negative).
To prevent either, the eigenvalues need to have zero real parts, i.e.
to be \emph{purely imaginary}.

The relationship between continuous time (as defined in the above
differential equation) and discrete time (as defined in a recurrence)
is explicit when taking steps of time $1$: 
\begin{equation}
x(t+1)=e^{M}x(t)\qquad\to\qquad x_{n+1}=Bx_{n}\qquad\mathrm{with}\quad B=e^{M}\label{eq:discrete}
\end{equation}
from where the relationship between continuous time evolution and
discrete time evolution is one of matrix exponentiation. Therefore,
in discrete time, the matrix property that is relevant to asymptotic
evolution is whether the eigenvalues of $B$ \emph{lie in the unit
circle}, i.e., have an absolute value of 1, because then their powers
do not explode or shrink. The exponential of imaginary numbers lies
in the unit circle. 

\subsection{Exponentials, Taylor, rotations}

The formal exponential of an operator $P$ is defined through the
series expansion
\[
e^{P}\equiv I+P+P^{2}/2+P^{3}/3!+P^{4}/4!+\cdots
\]
with $I$ the identity, and we will use this notation for matrices
and for other linear operators such as the derivative. Of course,
this definition might not converge into a well-defined operator. For
operators in finite-dimensional spaces with bounded eigenspectra it
will always formally converge. 

To give a simple example for matrices, considering the simplest antisymmetric
matrix, the 2x2 matrix 
\[
J=\left[\begin{array}{cc}
0 & 1\\
-1 & 0
\end{array}\right]
\]
we note that $J^{2}=-I$ and so formally operates as the imaginary
unit, the square root of $-1$. In particular 
\[
e^{Jt}=\left[\begin{array}{cc}
\cos(t) & \sin(t)\\
-\sin(t) & \cos(t)
\end{array}\right]=\cos(t)\,I+\sin(t)\,J
\]
so the exponential of a 2D antisymmetric matrix is a 2D rotation. 

A nontrivial example is given by the derivative operator. The Taylor
expansion embodies a translation operator $T_{\Delta}$, the linear
operator in function space that transforms $f(x)$ to $f(x+\Delta)$
:

\[
T_{\Delta}f(x)\equiv f(x+\Delta)=f(x)+\Delta f'(x)+\frac{\Delta^{2}}{2}f''(x)+\frac{\Delta^{3}}{3!}f'''(x)+\cdots
\]
and it is useful to remember that this translation operator can be
succintly written as a formal exponential of a derivative operator:
\[
f(x+\Delta)=e^{\Delta\frac{d}{dx}}f(x)\qquad\Rightarrow\qquad T_{\Delta}\equiv e^{\Delta\frac{d}{dx}}
\]
The spectrum of the derivative operator is not bounded, and therefore
there is no guarantee that this definition yields a well-defined operator.
In fact it does not: while the operator $T_{\Delta}$ is well defined
for any function, the operator $e^{\Delta\frac{d}{dx}}$ is only well-defined
when applied to functions that have a globally convergent Taylor series,
aka holomorphic functions (analytic functions without singularities,
such as the exponential).

In the theory of Lie groups {[}Hall{]}, groups of transformations
such as rotations and translations are called \emph{actions, }and
are obtained as the exponential of an \emph{infinitesimal generator}.
The infinitesimal generator of rotations are antisymmetric matrices,
and the infinitesimal generator of translations is the derivative
operator, or the gradient in $R^{N}$. 

\subsection{Orthogonal and unitary matrices }

A simple prescription to generate purely imaginary spectra is to create
an antisymmetric (aka skew-symmetric) matrix. To prescribe an antisymmetric
matrix you need only the upper triangle \emph{above} the diagonal,
and so the number of independent elements is $N(N-1)/2$. The other
elements are generating by reflection. Antisymmetric matrices are
normal, closed under addition and form an algebra called $so(n)$.
This simple prescription misses many matrices which do have purely
imaginary spectra but are not antisymmetric. 

The matrix exponentials of antisymmetric matrices are orthogonal matrices
with unit determinant; they have the structure of a Lie group, called
SO(n), and geometrically they form the Stieffel manifold. Their eigenvalues,
being the exponentials of the eigenvalues of the generating matrix,
are of the form $e^{i\theta}$ with unit absolute value. Such matrices
preserve norm, and their determinant is 1 so they preserve volumes.
Repeated multiplication by such matrices neither grows nor shrinks.
Orthogonal matrices contain all rotations in the $N(N-1)/2$ possible
rotation planes, all permutations of the axes, and various other operations.
They embody the set of linear transformations that leaves the unit
sphere in $N$ dimensions invariant. 

Generating orthogonal matrices is, in its full generality, computationally
nontrivial {[}Gallier,Cardoso{]}. Exponentiating an antisymmetric
matrix requires in general effort $O(N^{3})$ if calculated through
eigensystems, or $N^{2}M$ where $M$ is the number of terms in the
Taylor expansion guaranteeing numerical convergence. Padé approximant
methods and squaring rescaling can improve convergence {[}Cardoso{]}. 

Orthogonal matrices were used in {[}Saxe{]} and {[}Le{]} as initialization
on the network weigths. Both {[}Arjowski{]} and {[}Vorontsov{]} proposed
the use of orthogonal matrices and their complex-valued cousins, the
unitary matrices, as part of the ongoing dynamcis, to address the
vanishing gradient problem. In principle this works, but in practice
both came up against the same problem: making changes to the defining
$M$ in such a way as to preserve the orthogonality or unitarity of
the matrix is computationally expensive, a problem they called ``parametrization
of the Stiefel manifold''. {[}Arjowski{]} proposed a method based
on composition of a sequence of parametrized unitary transformations.
An alternative presented in {[}Chang{]} was to return to a network
defined by a differential equation rather than a recurrence, in which
case all that needs to be preserved is skew-symmetry of the synaptic
matrix. Such connectivities have also been explored from the neuroscience
side {[}Alonso,Magnasco{]}. However, a system defined through a differential
equation is relatively computationally inefficient as a numerical
integration method must be used to propagate forward. In fact the
proposal in {[}Alonso{]} was to move from ODE to recursion, and the
exponentiation required to go from ODE to Eq. \ref{eq:RNN} was derived. 

The complex-valued generalization of antisymmetry (skew-symmetry)
is anti-Hermitian matrices, which satisfy $M=-M^{\dagger}$ where
the $\dagger$ operation represents the complex-conjugate of the transpose.
AntiHermitian matrices have purely imaginary spectra. The exponential
of an anti-Hermitian matrix is unitary. A unitary matrix satisfies
$UU^{\dagger}=U^{\dagger}U=I$ and has eigenspectrum on the unit circle.
Like orthogonal matrices, they represent all rotations, and in addition
they contain all rotations of individual complex-valued elements around
the origin of the complex plane. 

\subsection{Convolutions with a kernel $K$ as the linear operator $K\otimes$ }

Many systems benefit from study using an architecture in which the
connections are ``translationally invariant'': a convolutional network
or convnet {[}LeCun1,2,3{]}, and this entire paper is about convolutions.
It is important to keep in mind that although the convolution kernel
appears to be a matrix, the relevant properties of the convolution
\emph{as a linear operator} are given by a \emph{much larger} matrix,
the asymptotic stability of the system is determined by the eigenvalues
of this \emph{much larger matrix}. In this paper I will talk about
the convolution $K\otimes X$ where $K$ is the kernel, $\otimes$
the convolution operator, $X$ the underlying layer; I will ruthlessly
abuse notation and call this larger matrix ``the linear operator
$K\otimes$''. 

To be clear, consider a layer $X$ which is a $1000\times1000$ image,
and we apply a Gaussian blur kernel $K$. Formally, the elements of
a Gaussian blur are never zero, but in practice they become negligibly
small outside of a circle of $8\sigma$ and in practice much smaller.
So, the \emph{nonzero core} of our Gaussian blur kernel could be an
$11\times11$ array, even though the kernel is formally $1000\times1000$
like the underlying layer. 

The convolution $K\otimes X$ is a linear operation, and therefore
\emph{must }be described by a matrix acting on the space of the layer.
For the purpose of linear algebra, the layer is a vector in a $1000000$-dimensional
space, since it is made out of $1000000$ numbers. A matrix acting
on this space has dimensions $1000000\times1000000$. How do we go
from the 121 elements of our kernel $K$ to a linear operator $K\otimes$
whose representation as a matrix has a trillion elements and one million
eigenvalues?

We need to think then of the canonical isomorphism between $\mathbb{R}^{1000\times1000}$,
the space we think of when we look at our layer $X$, and $\mathbb{R}^{1000000}$,
the space where linear algebra naturally lives, for example eigenvalues
and eigenvectors. The canonical mapping is to stack rows one after
each other (what Python does to store a matrix in the linear RAM)
or columns one after each other (what Fortran does), and is sometimes
called ``flattening'' or ``lift'', and is given by the ``reshape''
operator in Python; we'll denote flattening the layer $X\in\mathbb{R}^{1000\times1000}$
by a square bracket $\left[X\right]\in\mathbb{R}^{1000000}$. Our
$K\otimes$ operator is represented by a matrix, which we will denote
by $\left[K\otimes\right]$. Each one of the million rows of this
matrix multiply together all of the pixels in the flattend image by
the million entries in the row, only 121 of which are nonzero, and
assigns the result to one given pixel in the output corresponding
to the row. The 121 nonzero values are the same in every row of the
matrix, but shift around because the kernel gets moved, to be centered
on the output pixel. Most of these are around diagonals: the main
diagonal has the value of the center of the kernel, the first diagonal
the value of the kernel to the right of center, the 1000th diagonal
contains the value of the pixel above the center, etc. 

In the rest of this paper I will call a kernel $K$ symmetric, antisymmetric,
Hermitian, or any other matrix property, whenever the matrix $\left[K\otimes\right]$
has that property. For instance, flipping a kernel spatially through
its center pixel, causes its matrix to be transposed, so if a kernel
is symmetric under such a flip, the matrix is symmetric. 

A boring and obnoxious calculation shows that the rows of the square
of this matrix are given by convolving the rows of the matrix with
themselves \emph{as vectors, }and the rest of the calculation follows
easily. The boring details are in supplementary section S1. Thus I
can continue to obnoxiously abuse notation to note that
\[
\left[K\otimes\right]^{2}=\left[K\otimes\right]\times\left[K\otimes\right]=\left[(K\otimes K)\otimes\right]
\]
where $\times$ is the standard matrix product; that is, the matrix
obtained by squaring, as a matrix, the $\left[K\otimes\right]$ matrix
\emph{is the same} as the matrix corresponding to the kernel $K\otimes K$
(the right hand side), from where we can show that, \emph{as operators},
\[
(K\otimes)(K\otimes)=(K\otimes K)\otimes
\]
which is the functional composition of the linear operator $K\otimes$
with itself (the functional square) equals the linear operator associated
to the convolution of the kernel with itself. 

Because this can be applied recursively to $\left[K\otimes\right]^{N}$,
this allows me to drop the parenthesis everywhere in Equation \ref{eq:convex_formal}.
\emph{This entire paper is about the fact that these parentheses can
be dropped. }

\section{The convolutional exponential and its computation by FFT}

In this section we will derive the convolutional exponential operation
for the standard convolution architecture, which allows us to solve
in closed form the evolution of a system which is given by a convolutional
linear differential equation. 

Consider as a motivating example the following set of coupled differential
equations

\[
\dot{y}_{ij}=y_{i+1,j}+y_{i-1,j}+y_{ij+1}+y_{ij-1}-4y_{ij}\qquad\forall ij
\]
where the $y_{ij}$ are variables located on a square lattice whose
rows are $i$ and columns are $j$. This equation implements a finite-difference
scheme for the diffusion equation $\dot{y}=\triangle y$. The key
is that this equation takes the form of a convolution: the lattice
elements $y_{ij}$ are \emph{convolved} with a kernel of the form
\begin{equation}
\begin{array}{ccc}
 & 1\\
1 & -4 & 1\\
 & 1
\end{array}\label{eq:laplace}
\end{equation}
and the result of that convolution is then used as the dynamical law
for the differential equation. 

This differential equation admits a closed-form, analytic solution.
Calling the convolution kernel $K$ and the convolution operation
$\otimes$, we write the equation as 
\begin{equation}
\dot{y}=K\otimes y\label{eq:convODE}
\end{equation}
from where, by taking additional time derivatives, we can obtain $\ddot{y}=K\otimes K\otimes y$
and $\dddot{y}=K\otimes K\otimes K\otimes y$ and so forth. Using
the Taylor expansion $y(t)=y(0)+\dot{y}(0)t+\ddot{y}(0)\frac{t^{2}}{2!}+\dddot{y}(0)\frac{t^{3}}{3!}+\cdots$
, one reaches the expression 

\[
y(t)=\left[1+\frac{t^{1}}{1!}K+\frac{t^{2}}{2!}K\otimes K+\frac{t^{3}}{3!}K\otimes K\otimes K+\cdots\right]\otimes y(0)
\]
which defines an exponential operation for the convolution operator.
(An explicit proof of this is in Supplementary Materials S1). 

The convolution theorem proves that the Fourier transformation of
a convolution is an elementwise product. We can use this fact recursively
to note a convolution repeated $T$ times is, in Fourier space, the
$T$th power of the Fourier transform of the kernel. Calling $\mathscr{F}$
the (D-dimensional) Fourier transform and $\tilde{K}=\mathscr{F}\left[K\right]$
we first use 
\[
\mathscr{F}\left[K\otimes K\right]=\mathscr{F}\left[K\right]^{2}=\tilde{K}^{2}\implies\mathscr{F}\left[K\otimes K\otimes K\right]=\tilde{K}^{3}\implies\cdots
\]
where the powers of $\tilde{K}$ are taken \emph{pointwise}, to reach 

\[
y(t)=\mathscr{F}^{-1}\left[1+\frac{t^{1}}{1!}\tilde{K}+\frac{t^{2}}{2!}\tilde{K}^{2}+\frac{t^{3}}{3!}\tilde{K}^{3}+\cdots\right]\otimes y(0)
\]
and since all powers are elemenwise, each element resums to an \emph{elemenwise
exponential}. Therefore the analytic solution at time $t$ is obtained
as a convolution with a kernel which is given by 
\[
G\equiv\mathscr{F}^{-1}\left[1+\frac{t^{1}}{1!}\tilde{K}+\frac{t^{2}}{2!}\tilde{K}^{2}+\frac{t^{3}}{3!}\tilde{K}^{3}+\cdots\right]=\mathscr{F}^{-1}\left[\exp(t\mathscr{F}\left[K\right])\right]
\]
where the exponential operation is taken \emph{elementwise}. We will
call $G$ the \emph{convolutional exponential} of the kernel $K$,
and it embodies the full analytic solution for arbitrary times $\Delta.$
In technical dynamical systems jargon $G$ is a flow. In Physics,
the linearity of Eq. \ref{eq:convODE} leads to extensive use of $G$,
called a Green function for the equation. We will abuse notation and
write 
\[
e_{\otimes}^{K}\equiv\mathscr{F}^{-1}\left[\exp(\mathscr{F}\left[K\right])\right]
\]
to distinguish the convolutional exponential from either the pointwise
exponentiation or the matrix exponential. Finally, using the arguments
and notation introduced in Section 2.5, it follows that 
\[
\left[e_{\otimes}^{K}\otimes\right]=e^{\left[K\otimes\right]}
\]
or, in other words, the matrix representing the action of the convolutional
exponential of $K$ is, in fact, the matrix exponential of the matrix
representing the action of $K.$

Care must be taken, because the self-convolutions of a kernel make
its nonzero core grow in size; therefore in order to use FFT the kernel
needs to be zero-padded to a sufficient size to contain the convolved
kernels. This is the main limitation on performance. Since in principle
the size is limited by the size of the layer itself, and if the total
number of elements of the layer is $N$ then exponentiation is at
most $N\ln N$. If we compute this convolutional exponent on the kernel
\ref{eq:laplace}, we observe a spreading Gaussian of width $\sqrt{t}$,
see Figure FF. 

\includegraphics[width=12cm]{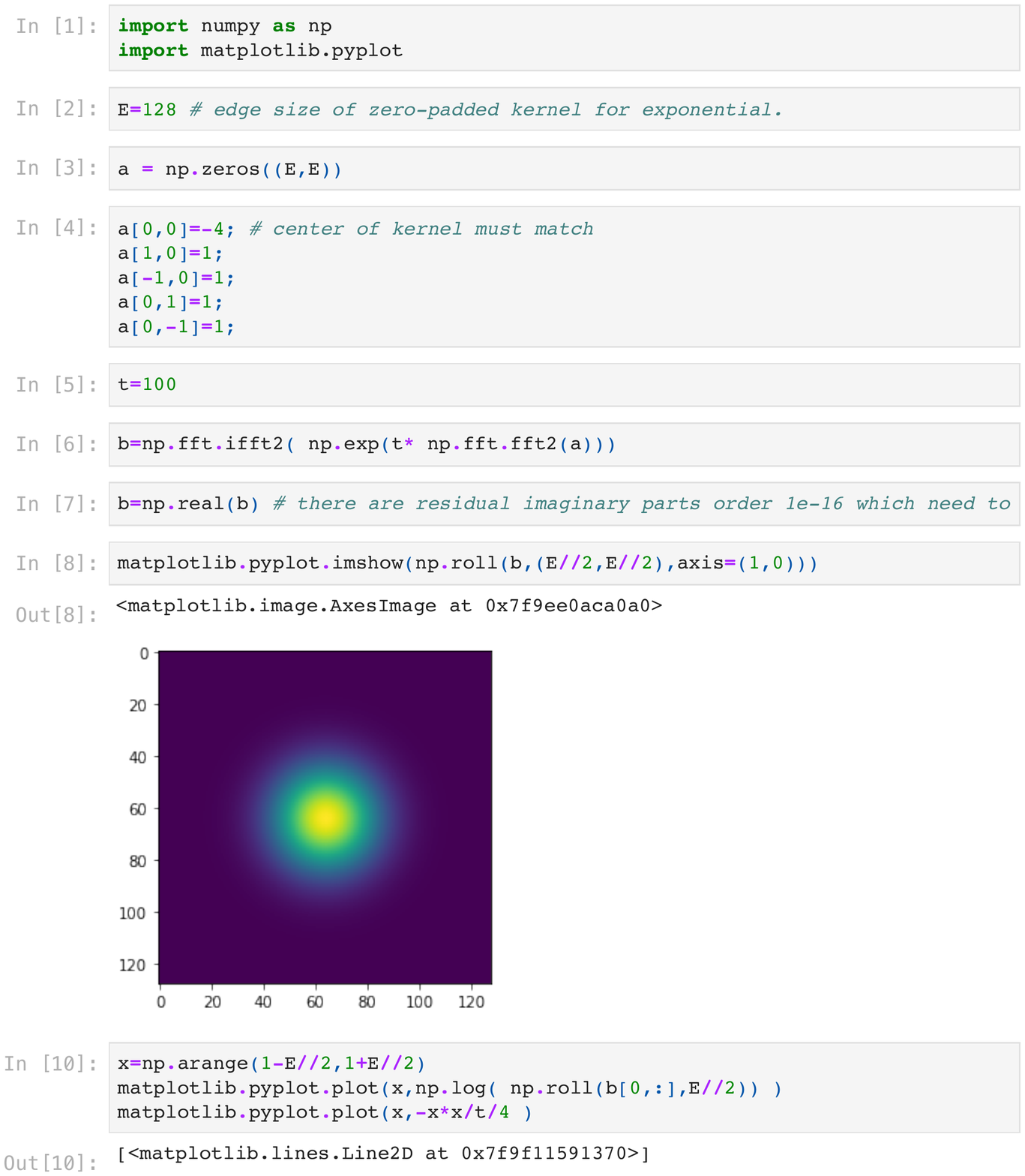}

Figure 1. Example code for exponentiating the Laplacian gives a 2D
Gaussian kernel of width $\sqrt{t}$. The full notebook with additional
demonstrations is in Supp. Mat. 
\begin{lyxcode}
\end{lyxcode}

\section{The convolutional unitary RNN}

Unitary complex-valued RNNs were introduced e.g. in {[}Arjovski{]},
where a full discussion of the architecture for the general case is
laid out. One way of generating a unitary matrix is by exponentiation
of a antiHermitian matrix, or equivalently, $i$ times a Hermitian
matrix. An anti-Hermitian matrix has two parts: the real part is antisymmetric,
and the imaginary part is symmetric. 

We want to extend this to convolutions. As discussed in section 2.5,
the kernel equivalent of the matrix transpose is a central symmetry
through the center of the kernel, i.e., swapping element $(i,j)$
with element $(-i,-j)$. Let us denote as $\intercal$ the complex
conjugate reflection through the center of symmetry, the kernel operation
corresponding to $\dagger$, the complex conjugate transpose of a
matrix. Then an antiHermitic kernel is one satisfying $K^{\intercal}=-K$.
For such a kernel, $e_{\otimes}^{K}$ is unitary, and then our dynamics
is 

\begin{equation}
Z_{n+1}=\phi\left(e_{\otimes}^{K}\otimes Z_{n}+I_{n}\right)\label{eq:cuRNN-2}
\end{equation}
with $\phi$ a complex valued activation function of a complex variable,
applied elementwise. This is a \emph{convolutional unitary} RNN. 

To generate an anti-Hermitian kernel we can start with any arbitrary
real kernel $U$ and extract its symmetric and antisymmetric components.
Then we can reassemble them, by multiplying the symmetric component
by $i$: 
\[
U\qquad\Leftrightarrow\qquad K=\left(\frac{U-U^{\intercal}}{2}\right)+i\left(\frac{U+U^{\intercal}}{2}\right)
\]
which is obviously a bijection. 

The result of exponentiating a matrix is, at least formally, invertible.
In practice, matrices having eigenvalues with large real parts are
going to have issues because upon exponentiation those components
become arbitrarily large or small. For example, exponentiating the
diffusion kernel for negative times gives rather ill-defined results,
and in the limit of large kernels diverges. This is the reason the
diffusion equation cannot be integrated backwards in time (it defines
a semiflow). However, exponentiating antiHermitian kernels has no
such issues, and \textbf{the resulting unitary kernels are readily
invertible}, allowing some measure of backtracking for invertible
activation functions. 

Finally, since the exponential is best computed in Fourier space,
we note that the Fourier transform of an antiHermitian kernel is purely
imaginary. Therefore, when each element is exponentiated, they acquire
unit absolute value for all coefficients. This is what is otherwise
called ``spectrally white''. 

\section{Derivatives of the convolutional exponential }

When the synaptic weights are not independent, but are generated by
a transformation, a number of training strategies require taking the
derivative of the synaptic weights with respect to the underlying
parameters used in the transformation. In our case, we generate unitary
kernels by exponentiation of an antiHermitian kernel; we will need
the derivative of the unitary kernel with respect to an arbitrary
element in the antiHermitian kernel. 

We will prove the derivative formula in 1D and leave the N-dimensional
case as an exercise. Given a 1D kernel $K$ whose coefficiets we note
as $K_{i}$, we want to compute the derivative of the exponential
kernel (itself a kernel) with respect to one specific coefficient
in the kernel in position $a$, to get a family of kernels $D^{a}$
parametrized by $a:$
\[
D^{a}=\frac{\partial}{\partial K_{a}}\exp_{\otimes}K
\]
which we can expand to
\[
D^{a}=\frac{\partial}{\partial K_{a}}\mathscr{F}^{-1}\left[\exp(\mathscr{F}\left[K\right])\right]=\mathscr{F}^{-1}\left[\frac{\partial}{\partial K_{a}}\exp(\mathscr{F}\left[K\right])\right]=\mathscr{F}^{-1}\left[\exp(\mathscr{F}\left[K\right])\frac{\partial\mathscr{F}\left[K\right]}{\partial K_{a}}\right]
\]
and since the Fourier transform is a matrix multiplication of the
input vector by the Fourier matrix $e^{2\pi i\frac{jk}{N}}$, the
derivative of this linear operation is simply the column $a$ of the
matrix
\[
\frac{\partial\mathscr{F}\left[K\right]_{j}}{\partial K_{a}}=e^{2\pi i\frac{ja}{N}}
\]
and this generates a \emph{translation} by $a$ (which in FFT space
is a circular shift by $a$): 
\[
\left(D^{a}\right)_{k}=\left(e_{\otimes}^{K}\right)_{k+a}
\]

The result holds in higher dimensions where the translation is along
multiple indices. 

A similar calculation shows the derivative of the convolutional sine
is the translated convolutional cosine, and the derivative of a cosine
is a translated convolutional sine. Therefore, once the original kernels
required for iteration of Eq. \ref{eq:coRNN} are computed, the derivatives
with respect to the elements are already at hand. 

\section{Antisymmetric convolutions}

The restriction of the above to the real numbers would generate orthogonal
kernels by exponentiation of antisymmetric kernels. As stated before,
for a convolution operation from a layer to itself to be antisymmetric
in the sense that the connections from $i\to j$ is minus the connection
from $j\to i$, the convolution kernel described must be antisymmetric
in \emph{space}, meaning flipping the kernel through its central location
switches the sign of the element. Such kernels, when exponentiated,
describe primarily translations and represent a large restriction
on possible connectivities. For example, a 1D kernel equal to $(-\frac{1}{2},0,\frac{1}{2})$
represents a derivative along $x$, and the exponential of such a
kernel generates finite translations along $x$, since the formal
exponential of a derivative translates into the Taylor expansion formula
describing a finite translation
\[
e^{\delta\frac{d}{dt}}=1+\delta\frac{d}{dt}+\frac{\delta^{2}}{2}\frac{d^{2}}{dt^{2}}+\frac{\delta^{3}}{3!}\frac{d^{3}}{dt^{3}}+\frac{\delta^{4}}{4!}\frac{d^{4}}{dt^{4}}+\cdots
\]

\[
e^{\delta\frac{d}{dt}}f(t)=f(t)+\delta f'(t)+\frac{\delta^{2}}{2!}f''(t)+\frac{\delta^{3}}{3!}f'''(t)+\cdots=f(t+\delta)
\]
or in Lie group jargon, the derivative is the infinitesimal generator
of finite translations {[}Hall{]}. It is easy to verify, for instance,
that adding to the diffusion kernel shown in Figure 1 a derivative-like
component, the exponential generates a translated Gaussian. (Supplementary
python notebook). 

Should straight antisymmetric convolutions be used, their exponentials
are given directly by the procedure of the previous section. Since
antisymmetric matrices have imaginary spectra, their exponentials
have spectra on the unit circle, and are therefore unitary. We have
found in practice that such a set of connections is not highly useful
by itself, but we will keep this in mind and return to it in a later
section. 

We proceed to derive a more general convolutional architecture with
a more generous parameter set. 

\section{Bipartite architecture}

A better way to generate a real-valued version of the convolutional
unitary RNN is, instead of exponentiating an antisymmetric kernel,
to exponentiate $i$ times a symmetric kernel. This would require
handling of real and imaginary parts but we can use a bipartite (symplectic)
trick to keep them separate. We make two copies of a system, which
we henceforth will be calling $X$ and $P$, then have arbitrary connections
$C$ from $X\to P$, and have all reciprocal connections from $P\to X$
be the negative value of the forward connection. This generates a
block structure in the overall connectivity matrix, if we first number
all elements of $X$ and then the homologous elements of $P$, where
the submatrix $C$ is \emph{arbitrary}: 
\[
\left[\begin{array}{cc}
0 & C\\
-C^{T} & 0
\end{array}\right]\left[\begin{array}{c}
X\\
P
\end{array}\right]
\]
Because the symplectic matrix 
\[
\left[\begin{array}{cc}
0 & I\\
-I & 0
\end{array}\right]
\]
is a square root of minus the identity, it functions formally as the
imaginary unit, and exponentials of matrices with such structure look
like rotation matrices. These can be supplemented by arbitrary intralayer
antisymmetric connections of course as seen in the previous section. 

\includegraphics[width=12cm]{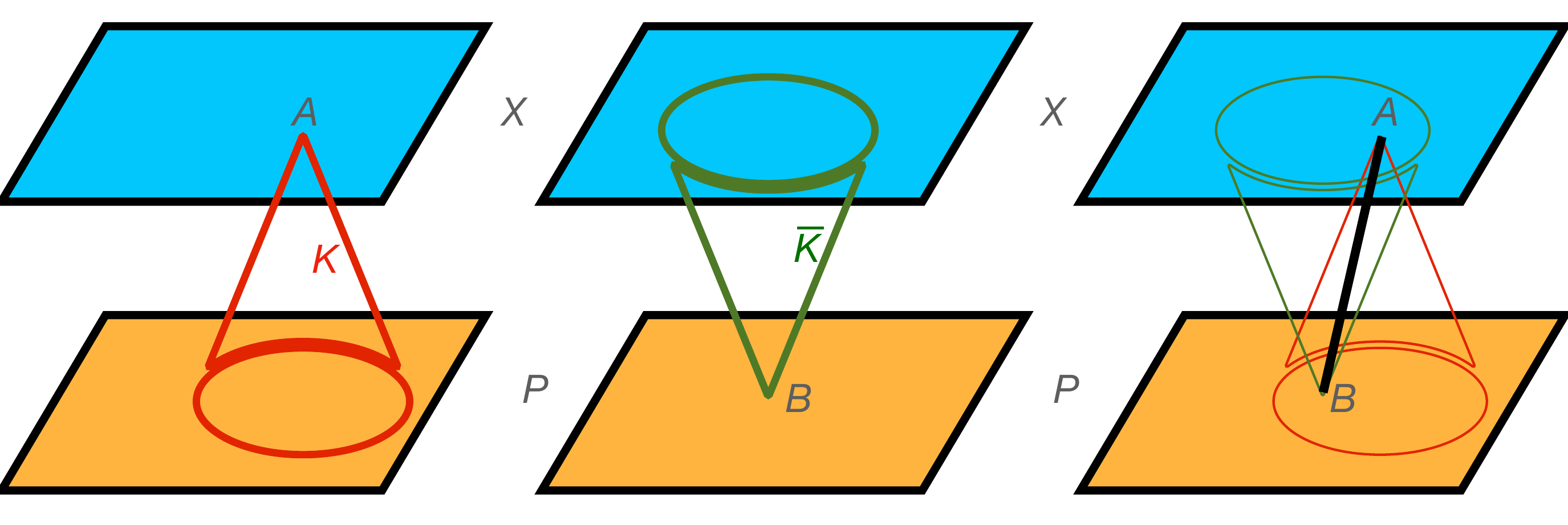}

Figure 2. Left: the (otherwise arbitrary) convolutional kernel $K$
integrates information from a subarea of $P$ to influence each element
$A$ of $X$. Middle: similarly the convolutional kernel $\tilde{K}$
does the corresponding thing in reverse. Right: for the connections
between $X$ and $P$ to be antisymmetric, in particular the connection
between $A$ and $B$ to reverse sign, the \emph{orientation} of the
kernel $\tilde{K}$ has to be reversed so that the same kernel element
that projects from $B\to A$ in $K$ is the one going from $A\to B$
in $\tilde{K}$. Given an arbitrary kernel $K$, the kernel $\tilde{K}$
computed with this prescription guarantees antisymmetry of the lifted
matrix. 

\section{Fast computation of bipartite convolutions}

If the connections from $X\to P$ are convolutional in nature (with
the convolution being in principle arbitrary), then as described in
Fig 2, to make the reciprocal connections antisymmetric two things
must be done: first the signs must be inverted. Second, the convolutional
kernel must be flipped along all axes (both horizontally and vertically
in 2D) so that the reciprocal element points \emph{back }at the original
element.

Similarly to section 2, we can derive the form for the convolutional
kernels by taking successive derivatives of the linear equation

\[
\dot{X}=K\otimes P
\]

\[
\dot{P}=-\tilde{K}\otimes X
\]
where $\tilde{K}$ is the kernel $K$ flipped in all directions as
indicated by Fig 2. Taking a second derivative

\[
\ddot{X}=-K\otimes\tilde{K}\otimes X
\]

\[
\ddot{P}=-\tilde{K}\otimes K\otimes P
\]
and a third derivative

\[
\dddot{X}=-K\otimes\tilde{K}\otimes K\otimes P
\]

\[
\dddot{P}=\tilde{K}\otimes K\otimes\tilde{K}\otimes X
\]
and the sin/cos structure of the composite exponential kernel starts
to develop: there will be 4 kernels, coupling $X,P$ to themselves
and each other. The self kernels will contain even powers while the
cross kernels will contain odd powers. Together with the alternating
sign structure these will have the power series of a cosine and a
sine. With all 4 kernels taken together, the convolutional operation
will be \emph{orthogonal} as a linear operator on the $(X,P)$ space,
and will preserve all volumes in this space. 

For the special case in which $\tilde{K}=K$ (a centrally symmetric
kernel) this structure is easy to see: 
\[
\left(\begin{array}{c}
X\\
P
\end{array}\right)(t)=\left(\begin{array}{cc}
+\cos_{\otimes}(tK)\otimes & +\sin_{\otimes}(tK)\otimes\\
-\sin_{\otimes}(tK)\otimes & +\cos_{\otimes}(tK)\otimes
\end{array}\right)\left(\begin{array}{c}
X\\
P
\end{array}\right)(0)
\]
where we have abused notation and nested the convolutions as a 2x2
matrix structure. The convolutional sines and cosines are given by
\[
\cos_{\otimes}(tK)=\mathscr{F}^{-1}\left[\cos(t\mathscr{F}\left[K\right])\right]
\]

\[
\sin_{\otimes}(tK)=\mathscr{F}^{-1}\left[\sin(t\mathscr{F}\left[K\right])\right]
\]

\section{Conclussions}

We have combined unitary and orthogonal evolution with convolutional
architecture, to explicitly obtain a convolutional unitary recurrent
network, and, using a simplectic trick, a convolutional orthogonal
recurrent network, for which all calculations required to parametrize
the orthogonal/unitary kernels are $N\ln N$, by exploiting the structure
of convolutions in Fourier space. In theory this would solve the vanishing
gradient problem. But as Yogi Berra remarked, \emph{in theory, theory
and practice are the same, but in practice they often aren't}. 

Orthogonal matrices and unitary matrices, by virtue of having eigenvalues
on the unit circle, are when iterated \emph{de facto} performing some
form of spectral analysis, using frequencies that can be \emph{anywhere}
on the unit circle, as opposed to a Fourier transformation which uses
evenly spaced frequencies. These frequencies, together with the spatial
shape of the kernel, are the targets of training; and so this family
of networks can be trained to discriminate rather complex and arbitrary
time dependencies; the convolutional nature of the network applies
this homogenously both in space as well as in time. We expect the
cuRNN/coRNN family to have applications where the input space is very
high-dimensional and time is quasi-continuous (multistream audio,
video). 

\section*{Acknowledgements}

The first ``scientific'' computer program I wrote in my career,
was to compute successive convolutions of probability distributions
using fft, on the PDP-11 of the La Plata Physics Dept, under the advice
of Oreste Piro. I hereby dedicate this paper to him. I would like
to thank Kateri Jochum and Mason Hargrave for help with the manuscript. 

\section*{Bibliography}

\textbf{So many references missing, please help me by pointing out! }

Alonso, Leandro M., and Marcelo O. Magnasco. \textquotedbl Complex
spatiotemporal behavior and coherent excitations in critically-coupled
chains of neural circuits.\textquotedbl{} Chaos: An Interdisciplinary
Journal of Nonlinear Science 28.9 (2018): 093102. 

Arjovsky, Martin, Amar Shah, and Yoshua Bengio (2016) \textquotedbl Unitary
evolution recurrent neural networks.\textquotedbl{} International
Conference on Machine Learning. PMLR, 2016. 

Cardoso, João R., and F. Silva Leite. \textquotedbl Exponentials
of skew-symmetric matrices and logarithms of orthogonal matrices.\textquotedbl{}
Journal of computational and applied mathematics 233.11 (2010): 2867-2875. 

Chang, Bo, et al. \textquotedbl AntisymmetricRNN: A dynamical system
view on recurrent neural networks.\textquotedbl{} arXiv preprint arXiv:1902.09689
(2019). 

Cook, Matthew (2004). \textquotedbl Universality in Elementary Cellular
Automata\textquotedbl{} . Complex Systems. 15: 1--40. 

Gallier, Jean, and Dianna Xu. \textquotedbl Computing exponentials
of skew-symmetric matrices and logarithms of orthogonal matrices.\textquotedbl{}
International Journal of Robotics and Automation 18.1 (2003): 10-20. 

Hall, Brian; Lie Groups, Lie Algebras, and Representations: An Elementary
Introduction (Graduate Texts in Mathematics, 222) 2nd ed. 2015, Corr.
2nd printing 2016 Edition

Hochreiter, S.; Bengio, Y.; Frasconi, P.; Schmidhuber, J. (2001).
\textquotedbl Gradient flow in recurrent nets: the difficulty of
learning long-term dependencies\textquotedbl . In Kremer, S. C.;
Kolen, J. F. (eds.). A Field Guide to Dynamical Recurrent Neural Networks.
IEEE Press. ISBN 0-7803-5369-2. 

Kaneko, Kunihiko. \textquotedbl Overview of coupled map lattices.\textquotedbl{}
Chaos: An Interdisciplinary Journal of Nonlinear Science 2.3 (1992):
279-282.

Kaneko, Kunihiko. \textquotedbl Theory and applications of coupled
map lattices.\textquotedbl{} Nonlinear science: theory and applications
(1993).

Kilian, Joe, and Hava T. Siegelmann. \textquotedbl The dynamic universality
of sigmoidal neural networks.\textquotedbl{} Information and computation
128.1 (1996): 48-56.

Le, Quoc V., Navdeep, Jaitly, and Hinton, Geoffrey E. A simple way
to initialize recurrent networks of rectified linear units. arXiv
preprint arXiv:1504.00941, 2015. 

LeCun, Yann, and Yoshua Bengio. \textquotedbl Convolutional networks
for images, speech, and time series.\textquotedbl{} The handbook of
brain theory and neural networks 3361.10 (1995): 1995. 

LeCun, Yann; Léon Bottou; Yoshua Bengio; Patrick Haffner (1998). \textquotedbl Gradient-based
learning applied to document recognition\textquotedbl{} (PDF). Proceedings
of the IEEE. 86 (11): 2278--2324. CiteSeerX 10.1.1.32.9552. doi:10.1109/5.726791. 

LeCun. Y, B. Boser, J. S. Denker, D. Henderson, R. E. Howard, W. Hubbard,
L. D. Jackel, Backpropagation Applied to Handwritten Zip Code Recognition;
AT\&T Bell Laboratories 

Magnasco, Marcelo O. \textquotedbl Robustness and Flexibility of
Neural Function through Dynamical Criticality.\textquotedbl{} Entropy
24.5 (2022): 591. 

Nair, Vinod and Geoffrey E Hinton. Rectified linear units improve
restricted boltzmann machines. In International conference on machine
learning, pages 807--814, 2010.

Pascanu, Razvan; Mikolov, Tomas; Bengio, Yoshua (21 November 2012).
\textquotedbl On the difficulty of training Recurrent Neural Networks\textquotedbl .
arXiv:1211.5063 

Saxe, Andrew M., McLelland, James L., and Ganguli, Surya. Exact solutions
to the nonlinear dynamics of learning in deep linear neural networks.
International Conference in Learning Representations, 2014. 

Siegelmann, Hava T., and Eduardo D. Sontag. \textquotedbl Turing
computability with neural nets.\textquotedbl{} Applied Mathematics
Letters 4.6 (1991): 77-80.

Strogatz, Steve; Nonlinear Dynamics and Chaos: With Applications to
Physics, Biology, Chemistry, and Engineering, Second Edition (Studies
in Nonlinearity) (2014)

Vorontsov, Eugene, et al. \textquotedbl On orthogonality and learning
recurrent networks with long term dependencies.\textquotedbl{} International
Conference on Machine Learning. PMLR, 2017. 

\pagebreak{}

\subsection*{S1: More details on convolutions vs. matrix exponentials }

Convolutional networks consist of connections which are replicated
identically over the networks' underlying space (usually a regular
lattice or array). For example, consider an $E\times E$ square lattice
on which variables $x_{ij}$ are defined, where $i$ is the row and
$j$ is the column. In order to define a single vector with a single
index, we use the canonical mapping flattening the indices by concatenating
the rows one after each other: 
\[
(i<E,j<E)\to k\equiv i+Ej<E^{2}
\]
which reshapes an $E\times E$ square onto a single vector of length
$N=E^{2}$. We'll occasionally call this mapping the ``lift''; it's
computational implementaiton is called ``reshape'' in Python. Then
the lifted variables would be assigned through this mapping, 
\[
X_{i+Ej}=x_{ij}
\]

Consider now a simple first-neighbor convolution in which, for every
point in the lattice, you add up the first neighbors with coefficients

\[
y_{ij}=ax_{ij}+bx_{i+1,j}+cx_{i-1,j}+dx_{ij+1}+ex_{ij-1}\qquad\forall ij
\]
 corresponding to a convolutional kernel of the form 

\[
...\begin{array}{ccccc}
0 & 0 & 0 & 0 & 0\\
0 & 0 & c & 0 & 0\\
0 & e & a & d & 0\\
0 & 0 & b & 0 & 0\\
0 & 0 & 0 & 0 & 0
\end{array}...
\]
(we use the convention the origin is on top left). We call $K$ the
edge size of the convolutional kernel, in this case 3. This convolutional
operation has a very simple expression in the lift, since the first
neighbours on the same row are separated by 1 and the first neighbors
on the same column are separated by $E$: 

\begin{equation}
\left(\begin{array}{cccccccc}
a & b &  & {E-1\atop ...} &  & d\\
c & a & b &  &  &  & d\\
 & c & a & b &  &  &  & d\\
 &  & c & a & b\\
 &  &  & c & a & b\\
e &  &  &  & c & a & b\\
 & e &  &  &  & c & a & b\\
 &  & e &  &  &  & c & ...
\end{array}\right)\label{eq:unrolled}
\end{equation}
so this is a (very sparse!) pentadiagonal matrix, of size $E^{2}\times E^{2}$
(so $E^{4}=N^{2}$ total elements) having a center diagonal identically
equal to $a$, a supradiagonal containing $b$s, a diagonal at $+N$
containing $d$s, etc; or in more compact code notation, 
\begin{lyxcode}
M=a{*}diag(ones(E\textasciicircum 2,1),0)~+

b{*}diag(ones(E\textasciicircum 2-1,1),1)~+

c{*}diag(ones(E\textasciicircum 2-1,1),-1)~+

d{*}diag(ones(E\textasciicircum 2-E,1),E)~+

e{*}diag(ones(E\textasciicircum 2-E,1),-E)
\end{lyxcode}
For simplicity, I have swept under the rug the details of how to wrap
the convolution around the edge of the lattice. The simplest way to
deal with this will be to wrap around plus one, i.e. when you come
off the right edge of the $E\times E$ square you return on the left
side, but one row lower. With such boundary conditions the wrapping
becomes extremely simple in terms of the circshift operator, because
now the diagonals when they come off one edge they return to the other
side. Detailed treatment of other boundary conditions will complicate
our treatment needlessly. 

So now we can lift the entire convolution

\[
y_{ij}=ax_{ij}+b_{i+1,j}+c_{i-1,j}+d_{ij+1}+e_{ij-1}\qquad\forall ij
\]
to 

\[
Y=MX
\]
where we will call a matrix $M$ of this form ``convolutional'',
and its main feature will be that the rows repeat with a rotation
(modulo, as explained, boundary conditions). The matrix $M$ is obviously
sparse, having only $K^{2}$ nonzero diagonals. 

We now come to the question of how to compute a matrix operation on
$M$ in a way that exploits its unique structure. For example, in
the theory of differential equations one would like to find solutions
to equations of the form 
\[
\dot{X}=MX
\]
which then require computation of the matrix exponential of $M$ defined
through its power series as 
\[
e^{M}\equiv I+M+M^{2}/2+M^{3}/6+M^{4}/24+\cdots
\]
where obviously the powers of $M$ are matrix products. The exponential
operation on a normal matrix is sometimes computed via its eigensystem;
on our matrix $M$ this brute force approach would take $\mathcal{O}(E^{6})$
operations. For a sparse matrix with very few diagonals it becomes
more practical to use the Taylor expansion and sparse matrix primitives. 

This poses an interesting issue of numerical precision. The matrix
exponential of a sparse matrix is usually not sparse; every element
can get to be nonzero if there is a ``path'' of nonzero connections
between elements. Let us consider for simplicity the case 
\[
a=-4;b=c=d=e=1
\]
This is again the Laplacian operator in 2D considered in the previous
section. The equation $\dot{X}=MX$ is therefore the diffusion equation
in 2D, and the solutions to this equation are of the form 
\[
G=e^{tM}
\]
 where $t$ is the elapsed time. Now, $G$ is also a convolutional
matrix; and if we ``unlift'' its coefficients by remapping the rows
of $G$ onto a 2D convolutional kernel, what we get is a 2D Gaussian
which spreads outwards with a width of $\sqrt{t}$. (Calculation left
as a useful exercise to the reader). Now such Gaussian is everywhere
nonzero, but can in practice be truncated to numerical accuracy whenever
the kernel elements are smaller than numerical precision. As a result,
for small values of $t$ we obtain a $G$ which is also sparse, but
whose number of nonzero diagonals increases with increasing value
of $t$. 

Even then, a sparse matrix multiplication scheme takes $\mathcal{O}(E^{4})$
operations to run times the number of terms in the series expansion
that need to be computed. A sparse method only takes account of the
sparse nature of the matrix and does not in any way use the convolutional
nature. 

In order to exploit the convolutional structure, the first thing to
notice is that when evaluating $M^{2}$, each row is composed of the
convolution of the row with itself, as a vector. Convolutions appear,
superficially, to be operations taking quadratic time; however a classical
technique uses the fact that the Fourier transform of a convolution
is an element-wise product; this allows the convolution to be computed
in Fourier space by using one Fast Fourier transform and one inverse,
and then element-wise squaring each element. In other words, if $x$
is a row of $M$ and $y$ a row of $M^{2}$, then 
\begin{lyxcode}
y=ifft(fft(x).\textasciicircum 2)~
\end{lyxcode}
which is an $\mathcal{O}(N\log N)$ operation. Applying this reasoning
to every single power in the power series allows us then to explicitly
re-sum the infinite series and observe that computation of each row
$z$ of the matrix exponential $e^{M}$ can be achieved as 
\begin{lyxcode}
z=ifft(~exp(~fft(x)~))
\end{lyxcode}
where exp is the element-wise exponentiation operation. This is, once
again, an $\mathcal{O}(N\log N)$ operation. (The only care to be
exercised is that the elements are complex). 

In fact, this method allows us to compute any analytic function $f$
expandable in Taylor series, applied to a convolutional matrix, as
the ifft of element-wise $f$ applied to the fft of the original vector.
This will be immensely useful for calculating the \emph{derivatives
}of the exponentiation operation with respect to the kernel elements
for backpropagation. 

Please note that previously we used the two-dimensional FFT on the
convolutional kernel laid out in 2D, while right now we used the one-dimensional
FFT on the convolutional kernel when lifted (unrolled, reshaped) onto
a 1D vector. These operations give rise to the same result and their
computational complexity is the same. The point of using the lift
is to explicity show the equivalence to a matrix exponential as usually
studied in the theory of dynamical systems. 

\section*{S2: Turing universality of convRNNs. }

Turing universality of generic RNNs (as the number of neurons $\to\infty$,
since a Turing machine requires a formally-infinite tape) has long
been established {[}Siegelmann,Kilian{]}; in some of the proofs the
``Turing machine tape'' involves using the coefficients of the network,
which become more and more as the network's size grows. Convolutional
RNNs do not have that freedom: the size of the kernel and the size
of the layer are different, and when the layer grows the kernel does
not grow in size. Therefore I will sketch a quick-and-dirty proof
specific to cRNNs. 

I will use an efficient shortcut. Cellular automata are, in a sense,
the discrete version of cRNNs, and the Rule110 2-state-3-symbol automaton
has been proved to be Turing universal {[}Cook{]}; in fact is arguable
the smallest or simplest system that has been shown to be Turing.
I will show how to embed a cellular automata dynamics in a cRNN. First,
we define a cRNN with continuous variables such that when the input
layer contains only 0s and 1s, the iteration generates the CA evolution.
Then it will be a matter of showing that the errors can be kept bounded,
namely, that a small amount of imprecision in the initial conditions,
or a small amount of noise added, does not grow to disrupt the equivalence
between the cRNN evolution in real variables vs. the discrete (by
definition noise-free); in other words, the embedding of the CA in
the cRNN must be shown to be stable in the sense of dynamical systems
theory. 

A cellular automata consist of discrete variables $Z$ (in our case
boolean) laid out in discrete space $x$ and discrete time $t$. A
logical function $\phi$ takes the neighborhood of a site, and uses
them to compute the value at that site at the next timestep. The smallest
family of CAs takes a 1D spatial lattice and the first neighbors to
each side of a site (3 symbols total)
\begin{equation}
Z_{x}^{t+1}=\phi(Z_{x-1}^{t},Z_{x}^{t},Z_{x+1}^{t})\label{eq:tCA}
\end{equation}
where $\phi$ (the ``rule'') is a Boolean-valued variable of 3 Boolean
arguments. There are 256 such functions, giving rise to fewer actually
distinct automata given symmetries (such as $1\Leftrightarrow0$ or
space inversion). Rules are numbered by the following method (the
Wolfram code for the automata): the output value for the input arguments
is listed for every combination of inputs in descending order, and
then is read-off as a binary digit. For Rule 110, 
\[
\begin{array}{cccccccccc}
{\rm Args} & 111 & 110 & 101 & 100 & 011 & 010 & 001 & 000\\
\phi & 0 & 1 & 1 & 0 & 1 & 1 & 1 & 0\\
 &  & 64 & 32 &  & 8 & 4 & 2 &  & \sum=110
\end{array}
\]
In order to embed this into a cRNN, we use a convolution kernel $C=\{4,2,1\}$
where the convolution is centered on the $2.$ Then when the convolution
kernel is applied to values $X_{n-1}X_{n}X_{n+1}$ the output is $4X_{n-1}+2X_{n}+X_{n+1}$.
This maps the eight possible combinations of arguments listed in the
Wolfram code, from left to right, to the values $(7,6,5,4,3,2,1,0)$.
Thus any activation function $\psi$ mapping the values 
\[
\psi(7,6,5,4,3,2,1,0)\to(0,1,1,0,1,1,1,0)
\]
defines a cRNN 
\begin{equation}
X^{t+1}=\psi(C\otimes X^{t})\label{eq:tcrnn}
\end{equation}
 where the $X$ are real, which, if started at an $X^{0}$ which has
values \emph{strictly equal} to $0$ or $1$, will forever evolve
the Rule110 automaton. 

In addition to show the dynamics Eq \ref{eq:tCA} is strictly embedded
in \ref{eq:tcrnn}, we would like to show the embedding to be \emph{stable.
}Imagine the initial state of the $X$ equals some boolean state plus
small perturbations $\delta$, $X_{x}^{t}=Z_{x}^{t}+\delta_{x}^{t}$.
We would like to know that if at time $t=0$ the $\delta_{x}^{0}$
are very small $\left|\delta_{x}^{0}\right|\ll1$, then the successive
evolution guarantees that $\left|\delta_{x}^{t}\right|<\frac{1}{2}\forall t$,
i.e. the $\delta$ never grow enough to change the value of the dynamics
and the $Z_{x}^{t}={\rm round}(X_{x}^{t})\ \forall t,x$. The conditions
on $\psi$ to guarantee this are extremely well-known from dynamical
systems theory; in particular the slope $\psi'$ at the integers $0\to7$
must be strictly smaller in absolute value than $1/7$ for linear
stability. We can do even better, and guarantee that the slope is
in fact 0 at every integer, in particular, that at $(0,4,7)$ the
function has a (potentially indifferent) local minimum and at (1,2,3,5,6)
it has a (potentially indifferent) local maximum. This will guarantee
\emph{superstability, }in which a small deviation is rapidly quenched.
For example, at a locally quadratic maximum, a small deviation $\delta$
is mapped to $a\delta^{2}$ which\emph{ }regardless of the value of
$a$ is always much smaller than $\delta$ for sufficiently small
values. 

Playing with the values of the convolution kernel we may arrange the
target so that $\psi$ only need to be unimodal. For instance, using
$C=(2,2,1)$ and 
\[
\psi(x)=\frac{1}{1+e^{\sigma(x-0.5)}}\frac{e^{3\sigma}}{1+e^{\sigma(3.5-x)}}
\]
provides an extremely robust stable embedding of Rule 110 for $\sigma>15$,
immune to both small deviations in initial conditions and small additive
noise in the dynamics. 
\end{document}